# Generating retinal flow maps from structural optical coherence tomography with artificial intelligence

Cecilia S. Lee MD MS,[1] Ariel J. Tyring MD,[1] Yue Wu PhD,[1] Sa Xiao PhD,[1] Ariel S. Rokem PhD,[2] Nicolaas P. Deruyter BS,[1] Qinqin Zhang PhD,[3] Adnan Tufail MD FRCOphth,[4] Ruikang K. Wang PhD,[1,3] Aaron Y. Lee MD MSCI[1,2]

[1] Department of Ophthalmology, University of Washington, Seattle WA

[2] eScience Institute, University of Washington, Seattle WA

[3] Department of Bioengineering, University of Washington, Seattle WA

[4] Moorfields Eye Hospital NHS Foundation Trust, London UK

Corresponding author:

Aaron Y. Lee
Assistant Professor
Department of Ophthalmology
University of Washington
Box 359608, 325 Ninth Avenue
Seattle WA 98104
Ph: (206) 543-7250
Email: leeay@uw.edu

ABSTRACT

Despite advances in artificial intelligence (AI), its application in medical imaging has been burdened and limited by expert-generated labels. We used images from optical coherence tomography angiography (OCTA), a relatively new imaging modality that measures retinal blood flow, to train an AI algorithm to generate flow maps from standard optical coherence tomography (OCT) images, exceeding the ability and bypassing the need for expert labeling. Deep learning was able to infer flow from single structural OCT images with similar fidelity to OCTA and significantly better than expert clinicians ($P < 0.00001$). Our model allows generating flow maps from large volumes of previously collected OCT data in existing clinical trials and clinical practice. This finding demonstrates a novel application of AI to medical imaging, whereby subtle regularities between different modalities are used to image the same body part and AI is used to generate detailed inferences of tissue function from structure imaging.

INTRODUCTION

Advances in image analysis and artificial intelligence (AI) have made computer-aided diagnosis (CAD) widely applicable. However, supervised machine learning usually requires a large number of expert-defined labels, with two main limitations.(Valizadegan, Nguyen, and Hauskrecht 2013) First, most of these labels are manually generated by clinicians, which is a cumbersome, time-consuming, and consequently costly process. Second, the use of human generated annotations as the ground truth limits the learning ability of the AI, given that it is problematic for AI to surpass the accuracy of humans, by definition. In addition, expert-generated labels suffer from inherent inter-rater variability, thereby limiting the accuracy of the AI to at most variable human discriminative abilities. Thus, the use of more accurate, objectively-generated annotations would be a key advance in machine learning algorithms in diverse areas of medicine.

Optical coherence tomography (OCT) is a non-invasive imaging modality of structural retina *in vivo*. Since its development in 1991, OCT has become essential in diagnosing and assessing most vision-threatening conditions in ophthalmology.(Murthy et al. 2016) A recent advance in OCT technology led to its counterpart, OCT angiography (OCTA), which measures blood flow in retinal microvasculature by obtaining repeated measurements of phase and intensity at the same scanning position.(Wang et al. 2007; de Carlo et al. 2015) While OCTA can theoretically be obtained using the same OCT hardware, in practice, OCTA requires both hardware and software modifications to existing OCT machines. OCTA can visualize both superficial and deep capillary plexus of the retinal vasculature without an exogenous dye, unlike fluorescein angiography, enabling better detection of overall retinal flow without potential side effects.(Ting et al. 2017) Despite the advantages, the use of OCTA is not as widespread as OCT, due to its cost and limited field of view (FOV) on currently commercially available devices, which decreases the ability to assess microvascular complications of retinal vascular diseases. In addition, OCTA requires multiple acquisitions in the same anatomic location, limiting the ability to acquire interpretable images in eyes with unstable visual fixation and motion artifacts from microsaccades.(Q. Zhang et al. 2016)

Given the relationship of OCT and OCTA, we sought to explore the deep learning's ability to first infer between structure and function, then generate an OCTA-like en-face image from structural OCT image alone. Human graders struggle to identify all but the largest vessels on structural OCT scans. A successful model would result in the acquisition of new information from preexisting databases given the ubiquitous use of OCT and may result in en-face images significantly less affected by artifacts. Unlike current AI models which are primarily targeted towards classification or segmentation of images, to our knowledge, this is the first application of artificial neural networks to generate a new image based on a different imaging modality data. In addition, this is the first example in medical imaging to our knowledge where expert annotations for training deep learning models are bypassed by using objective, physiologic measurements.

RESULTS

Four different model archetypes (Figure 1A) were designed to take a single individual structural B scan image as input and provide an inferred flow B scan image as output which included 5 blocks of max pooling and upsampling with 5 convolutional filters, 5 blocks with 10 convolutional filters, 9 blocks with 9 convolutional filters, and 9 blocks with 18 convolutional filters. In addition for each set of the 4 models, 3 different bridge connections were tested: no bridge (similar to traditional convolutional autoencoder network), element-wise summation, and copy + concatenation. Each of these models were trained from random initialization with the same batch number, training/validation datasets, optimizer, and learning rate for 5,000 iterations (Figure 1B) and the deepest model with 18 convolution filters and with copy + concatenation bridges had the lowest MSE (Figure 1C). The final model had a total of 7.85 million trainable parameters and has a space complexity of 90 megabytes. The model received no additional information regarding the neighboring slices. An extended training session was performed with dropout layers for regularization (Figure 1D). After collecting independently inferred flow B scan images, an en-face projection image was created using the same techniques as OCTA.

A total of 401,098 cross-sectional structural, macular spectral-domain (SD) OCT images from 873 volumes were designated as a training set: these were presented to the deep learning model, and the output was compared to the corresponding retina-segmented OCTA image. Another 76,928 OCT images from independent 171 cubes were used for cross-validation against OCTA images. A held out test set of 92,606 images of 202 cubes from a different set of patients was then used for comparison of deep learning model performance against the OCTA images.

The model was trained with 60,000 iterations. The learning curve with the mean squared error (MSE) of the training iterations and the validation set are shown in Figure 1B. The model achieved a minimal cross-validation MSE of $9.9482 \times 10^{-4}$. The weights that produced the best cross-validation MSE were then used for comparison against OCTA. The performance on the held-out test set achieved an MSE of $7.7665 \times 10^{-4}$. The fidelity by peak signal to noise ratio was 31.10 db. Figure 2 shows examples of deep learning inference of retinal flow from cross-sectional structural OCT images of the held out test set compared to the corresponding OCTA images. From structural OCT images, the deep learning model was able to identify both the large and medium-sized retinal vessels as well as the retinal microvasculature at a level of detail similar to the OCTA image. In addition, the model identified small vessels that are not apparent on the structural OCT image. Furthermore, the model was able to learn the segmentation of the retina and isolate structural features of the retina. Given that the model takes as input only a single structural B scan image, we asked three masked independent retina specialty trained experts to identify vessels on a single structural B scan image. Comparison of model output to three different masked clinicians revealed that the trained model was able to significantly outperform clinicians in terms of specificity, positive predictive value, and negative predictive value when using OCTA as ground truth.

To examine the performance of the model on other retinal pathologies outside of what the algorithm was trained, the weights from the lowest cross validation MSE were used to infer flow of each cross-sectional structural OCT image in a volume. Average projection of resulting inferred flow volume was used to create en-face projection maps of flow (Figure 3). Surprisingly, even without three-dimensional knowledge of the location of vessels nor knowledge of the neighboring cross sectional slice, the inferred flow by deep learning generated contiguous vessel maps similar to OCTA.

Compared to en-face projections of the structural OCT volumes (Figure 3 A, D, G, J), the AI generated flow maps (Figure 3 B, E, H, K) show more detail of the superficial retinal vasculature. The map of retinal vasculature generated by the model was much more detailed at the superficial retina than deeper in the retina and was superior to structural OCT en-face projections. This discrepancy was easily demonstrated in the pathologic eyes, (Figure 3 E, H, K) in which the superficial capillary plexus were affected by ischemia more than deep capillary plexus. In eyes with diabetic ischemia (Figure 3 E, F) and branch retinal vascular occlusion (Figure 3 H,I), OCTA reveals a higher density of deep capillary plexus than our model's flow output. Intact cilioretinal artery in the setting of central retinal artery occlusion (CRAO) preserves the superficial capillary plexus in the area perfused by cilioretinal artery. As expected, our model's results were comparable to OCTA in the area perfused by cilioretinal artery but not able to show the remaining deep capillary plexus elsewhere in the macula. (Figure 3 K, L).

When compared to color fundus photography and en-face projections of structural OCT volumes, the retinal vasculature map generated by the model was better for visualizing the superficial capillary networks. As shown in Figure 4, the deep learning image highlights the area of capillary dropout and intact flow. Both superficial arterioles and superficial capillary networks are clearly visible in the area supplied by cilioretinal artery on deep learning image. In contrast, only larger retinal vessels are visible, and the integrities of capillary plexus are difficult to assess on the color photo and structural OCT en-face projections. Similarly, Figure 5 shows that, for a normal eye, the retinal vasculature map generated by the model demonstrates the superficial capillary networks with superior detail compared to both the corresponding color fundus photograph and the late phase fluorescein angiogram. OCTA continues to highlight the retinal microvasculature with the highest detail. Manual segmentation of vessels on all four imaging modalities was performed (Figure 6). For second order vessels, 87.5%, 97.5%, and 100% of the vessels were identified by color, FA, and deep learning generated flow maps. Deep learning was able to identify significantly more third order vessels compared to color images ($p = 0.0320$) and was able to identify significantly more fourth order vessels compared to both color and FA images ($p = 1.86 \times 10^{-5}$ and $5.01 \times 10^{-4}$ respectively).

DISCUSSION

Our study demonstrates that a deep learning model can be trained to recognize features of OCT images that allow successful identification of retinal vasculature on cross sectional OCT scans in a fully automated fashion and generate en-face projection flow maps. The deep learning model identified both the retinal vessels that were easily seen on structural OCT b-scan images as well as the retinal microvasculature that was not apparent to the clinician on standard OCT (Figure 2) and showed significantly more retinal vessels compared to structural OCT en-face projections, color and FA images. Surprisingly deep learning was able to generate detailed flow maps of the retinal vessels in a variety of retinal conditions using standard, ubiquitously available structural imaging.

Taking advantage of an already existing imaging modality as the ground truth, our study bypasses the need to generate expert annotations entirely. Furthermore, acquiring a retinal vascular map from OCT via deep learning enables the comparison of retinal vascular structure versus function of retinal vasculature using OCT and OCTA, respectively. One of the potential mechanisms by which deep learning infers flow from structural images may be similar to speckle-variance (SV) processing method. SV imaging measures decorrelation between the OCT signals that are generated by speckle or backscattered light from biological tissues.(Mahmud et al. 2013; A. Zhang et al. 2015) Different speckle patterns are created due to moving particles in biological tissues such as red blood cells, thus enabling measuring flow. However SV imaging requires repeated imaging in the same anatomic location, whereas the algorithm presented here is able to infer flow from a single structural OCT scan. Thus, deep learning may be detecting the likely decorrelation that exists on a single scan based on trained relationship between OCT and OCTA.

The novel application of deep learning in our study infers flow from traditional OCT images. This finding has significant clinical applications. First, OCT is the most commonly performed eye procedure, thus resulting in extensive OCT databases in most clinics (including those acquired before OCTA was available). This large imaging cohort may allow us to determine the natural history of vascular changes, blood flow, and clinical outcomes in retinal diseases, similar to previous studies involving fundus photographs(R. Klein et al. 2004, 2007) but allowing much precise 3D volumetric information of retinal vasculature and thickness of retinal layers. Second, we do not know whether the output of our deep learning algorithm is structural or functional information given that the input was strictly based on structural images but the training was performed with functional data (OCTA). If the algorithm is indeed providing only structural map of the vasculature, then the results may allow us to compare the discrepancy between structure and function when we compare them to OCTA images. With longitudinal data, we may discern when the structural changes occurs: prior to the decline in function or vice versa. Future comparisons of the OCTA and deep learning images of the eyes that underwent a recent vascular insult, in which a clear difference exists between structure and function would be useful. More research is needed to establish the utility of applying deep learning to structurally correlated images, but a similar principle could be applied in different fields of medicine where structural imaging is routinely obtained and functional imaging data is available for use of the ground truth, such as computerized tomography or magnetic resonance imaging.

In comparison to structural OCT en-face projections, we show that the deep learning inferred flow maps are able to provide better definition of retinal vessels (Figures 3, 4). Powner et al. has previously shown that en-face projections of the structural OCT volumes leave behind basement membranes which are unperfused but will appear hyperreflective on the structural OCT imaging.(Powner et al. 2016) This suggests that the AI generated flow maps may have more clinical utility than en-face projections of structural OCT volumes since the latter can not distinguish between perfused and unperfused vessels.

In our work, we utilized a U shaped autoencoder network as the final model that has traditionally been used for semantic medical segmentation.(Ronneberger, Fischer, and Brox 2015) This deep neural network architecture uses bridges to maintain high resolution spatial information that is normally lost during pooling operations. In addition, deeper neural networks have generally been found to improve performance which has been shown in the computer vision research with improved accuracy with ImageNet image classification as networks became deeper.(Simonyan and Zisserman 2014; He et al. 2015; Szegedy et al. 2015) With our data, we have empirically found that the copy and concatenation bridge with deeper number of convolutional layers led to the best performance compared to models with shallower networks, no bridge connections, and element-wise summation bridge connections. Future work could include further optimization and hyperparameter model selection and comparison to other architectures such as V-net,(Milletari, Navab, and Ahmadi 2016) recurrent convolutional layers,(Liang and Hu 2015) two-pathway convolutional networks,(Havaei et al. 2017) and hybrid models with convolutional layers with recurrent network layers.

More broadly, a similar approach could have application in many other imaging modalities, where the same object is imaged with sensitivity to different properties of the object that is being imaged. For example, radiological/MRI measurements of the same body part are routinely conducted with different contrasts, taking advantage of the sensitivity of different contrasts to different properties of the tissue. While this allows measurements of different tissue properties, it is also time-consuming. In some cases, measurements that are highly sensitive to certain tissue properties may require invasive injection of contrast agents, or exposure to x-ray radiation. The present results demonstrate that even a very small amount of sensitivity to variations in a tissue property may be enough for a deep neural network algorithm to detect variations in the dependent image properties, enabling accurate inference of the physiological features of the imaged body part, even when these features are not readily visible to an expert, and undetectable by means of other image processing algorithms. For example, MR angiography (MRA) and anatomical MRI are often both required for imaging of soft tissue and imaging of blood vessels in the same organ. If the anatomical MRI possesses subtle sensitivity to the structure of the blood vessels, as demonstrated in the present study for OCT, it is possible that information analogous that derived from MRA could be inferred directly from an anatomical MRI scan.

Our approach has a number of limitations. The data collected for this study was done at a single academic center with a device from a single manufacturer and a consistent imaging protocol. While this may limit the immediate generalizability of this method, the weights learned in this work could be used as the starting point for transfer learning,(Caruana 1995; Bengio and Others 2012) allowing the model to learn to infer with images from other devices rapidly, and with substantially less data. Additionally, future studies will need to evaluate the clinical correlations between our deep learning inferred flow maps and retinal perfusion which would include oxygenation of vascularized tissue, tissue oxygen consumption, and/or flow velocity.

In conclusion, we show that deep learning is able to generate flow maps of superficial retinal circulation using structural OCT images alone. This approach may be used to analyze existing OCT datasets or be integrated into existing OCT machines today. In addition, this methodology of inferring weakly correlated images may be useful in many other imaging applications.

AUTHOR CONTRIBUTIONS

Conception and design (CSL, AYL, AT); analysis and interpretation (CSL, AJT, YW, SX, ASR, AT, RKW, AYL); writing the article (CSL, AJT, AYL); critical revision of the article (YW, SX, ASR, NPD, QZ, AT, RKW); final approval of the article (CSL, AJT, YW, SX, ASR, NPD, QZ, AT, RKW, AYL); data collection (CSL, QZ, RKW, AYL); statistical expertise (CSL, AYL); literature search (CSL, AJT, NPD).

CONFLICT OF INTEREST DISCLOSURES

Ruikang Wang received research support from Carl Zeiss Meditec, Inc. and co-owns a patent of OCTA technology with Oregon Health & Science University. No other conflicts of interest exist for the remaining authors.

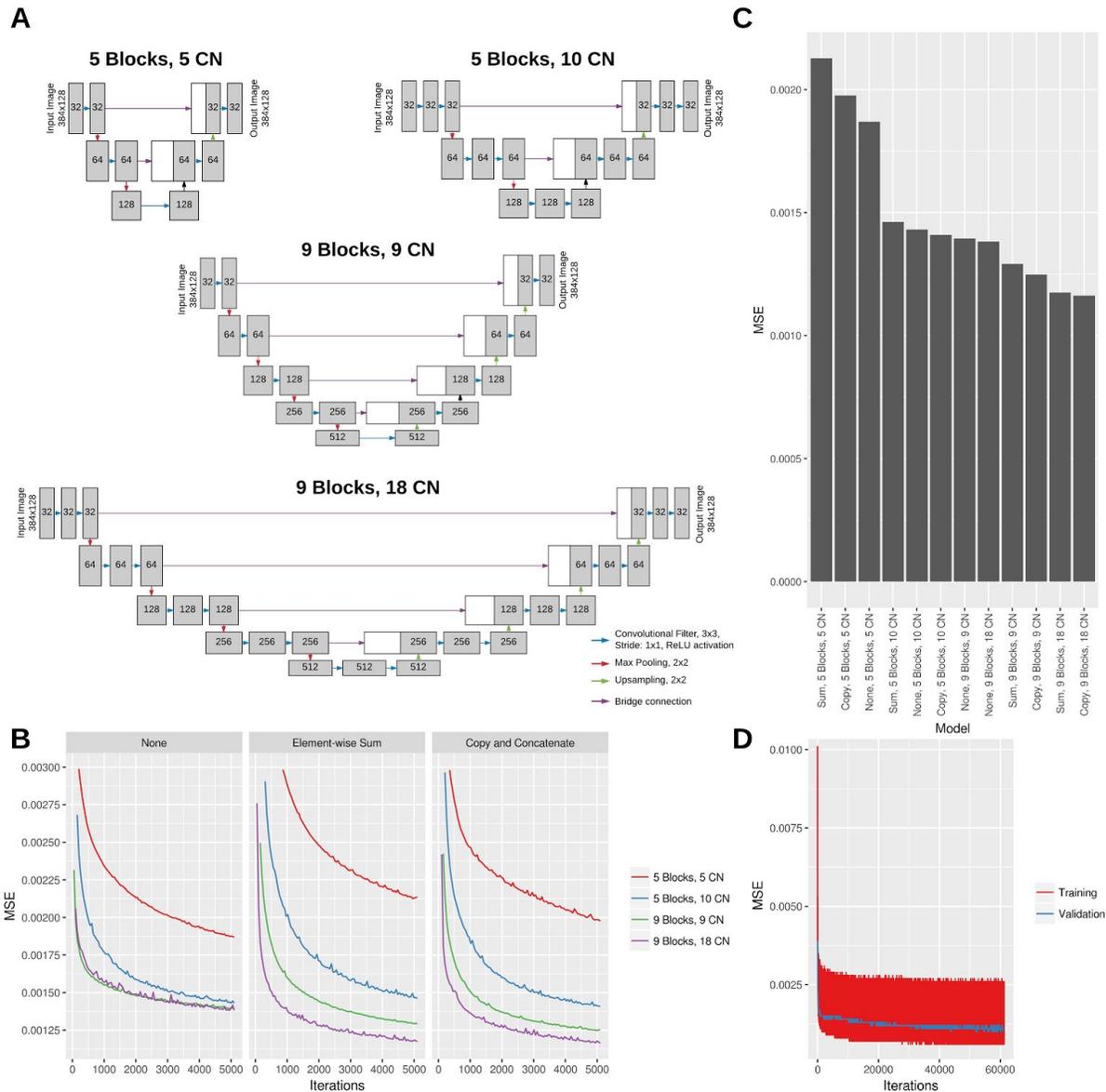

**Figure 1.** Schematics of deep learning model architectures tested by varying block depth and number of convolutional (CN) filters (A). Learning curve for all models with Mean Standard Error (MSE) for validation sets with the same batch size and learning rate and faceted by bridge type (B). Lowest MSE after 5,000 iterations for all models (C). Learning curves for the best performing (9 blocks, 18 convolutional filters, copy and concatenation bridge) on training and validation set for an extended training session (D).

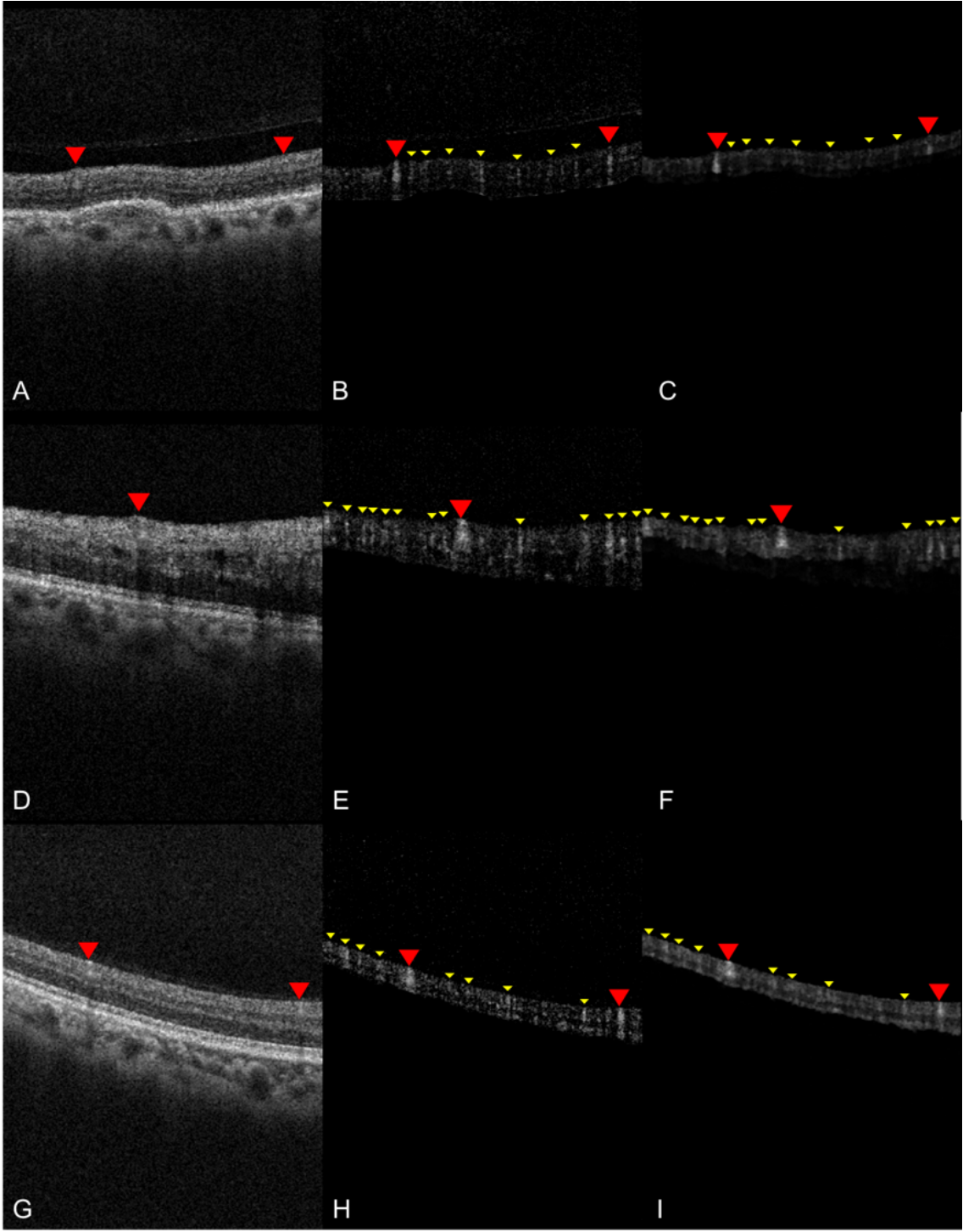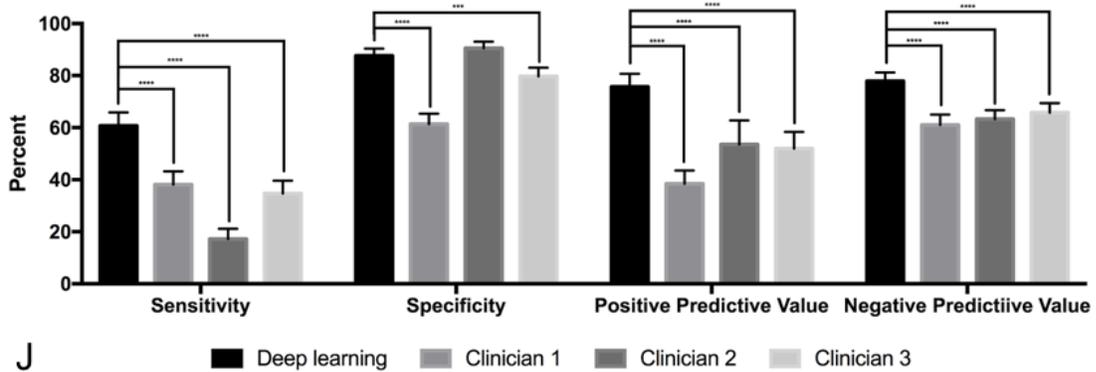

Figure 2: A,D,G. Example structural OCT images from held-out test set which serves as input for the deep learning model. The red arrowheads point to a large retinal vessel that is easily identified by expert clinicians. B,E,H. Example cross-sectional OCTA images from held-out test set. The red arrowheads point to the large retinal vessels and the yellow arrowheads point to the small hyperreflective areas that represent the retinal microvasculature and are not apparent on the structural OCT image. C,F,I. Example cross-sectional images of CNN output from held-out test set identifying retinal vessels. The deep learning model identifies both the large retinal vessels (red arrowheads) and small (yellow arrowheads) microvasculature (yellow arrowheads) similar to the OCTA images. J. Comparison of deep learning model against three masked retina-trained clinicians using OCTA as reference. *** P < 0.0001, **** P < 0.00001.

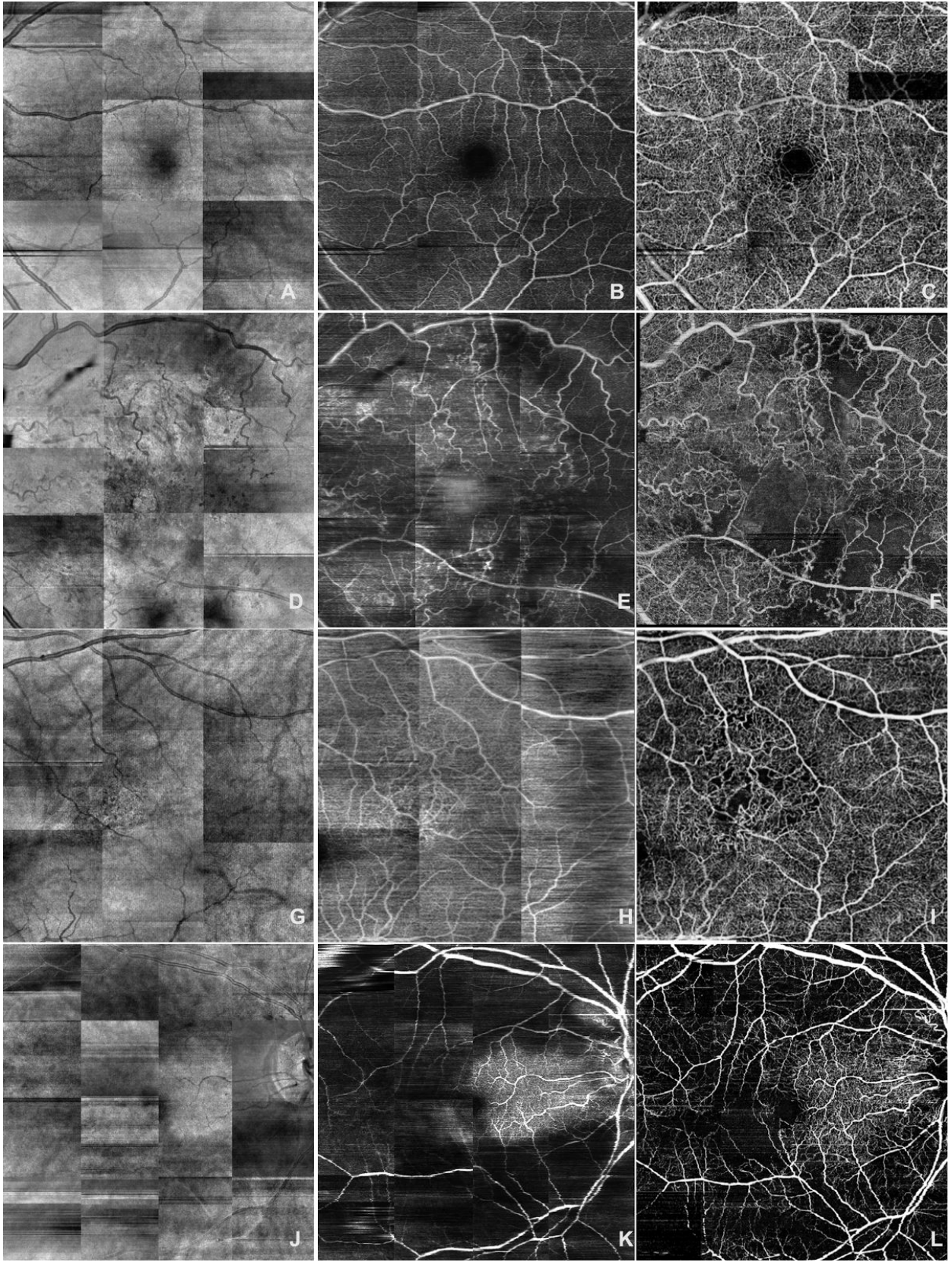

Figure 3: En-face projection maps of retinal blood flow created from structural OCT volumes (A, D, G, J), AI-generated inferred flow volumes (B, E, H, K) and OCTA flow volumes (C, F, I, L). Deep learning images demonstrate similar details of superficial retinal vasculature as the OCTA image while missing higher density of deep capillary networks which are visible on OCTA images. A, B, C Normal retina. D, E, F Diabetic retinopathy. G, H, I Branch retinal vein occlusion. J, K, L Central retinal artery occlusion.

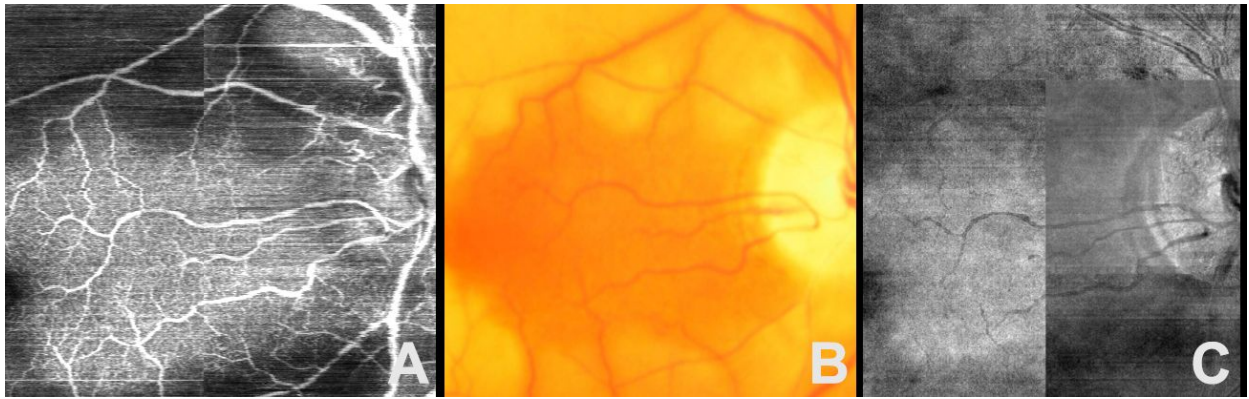

Figure 4. En-face projection maps of retinal flow created from inferred flow by deep learning (A) compared to the corresponding color fundus photo (B) and structural OCT en-face projection (C) of an eye with central retinal artery occlusion and intact cilioretinal artery.

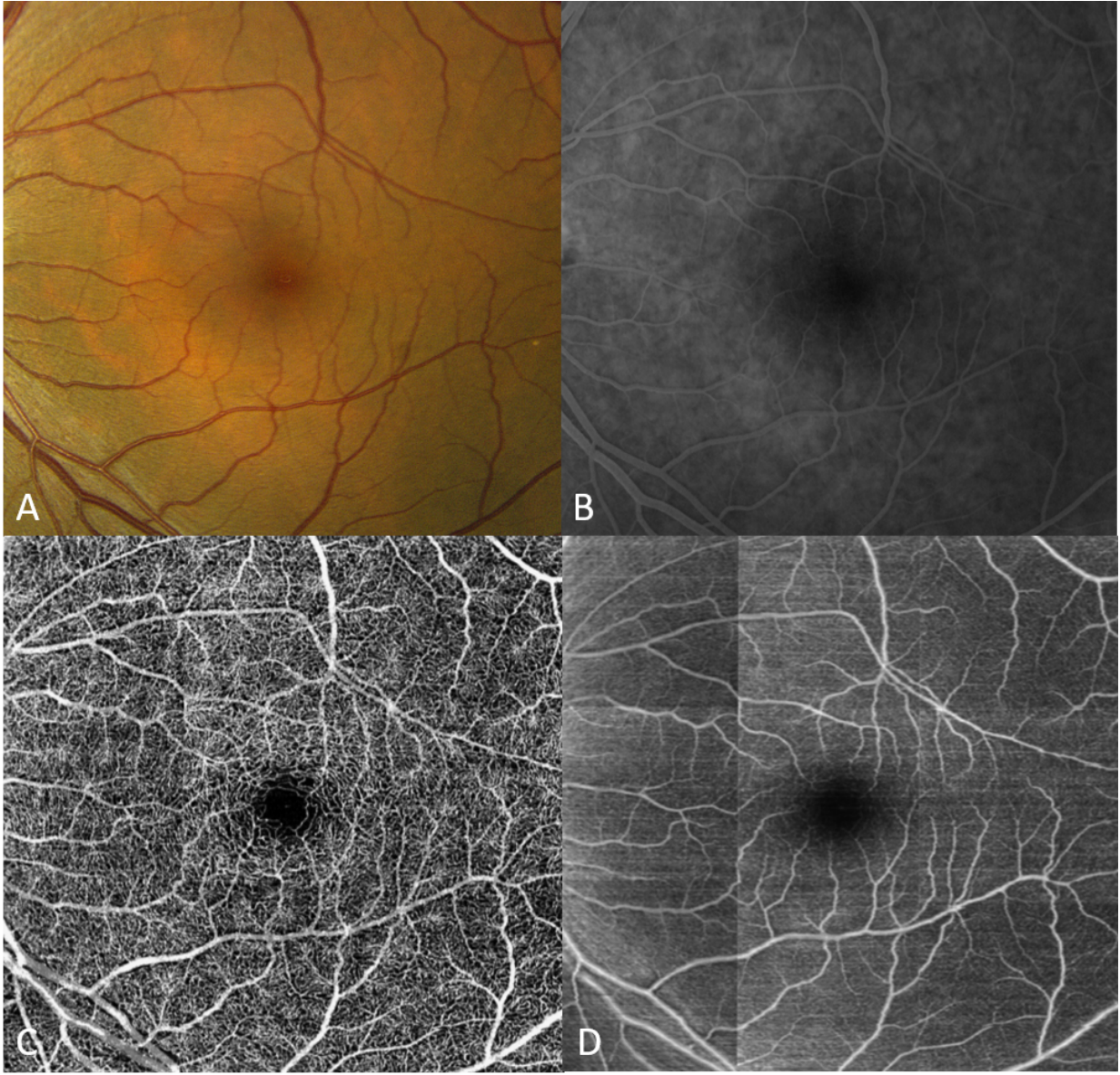

Figure 5. Color fundus photo (A) of a normal retina compared to the corresponding late phase fluorescein angiogram (B), the corresponding OCTA image (C), and contiguous vessel map generated by the deep learning model (D).

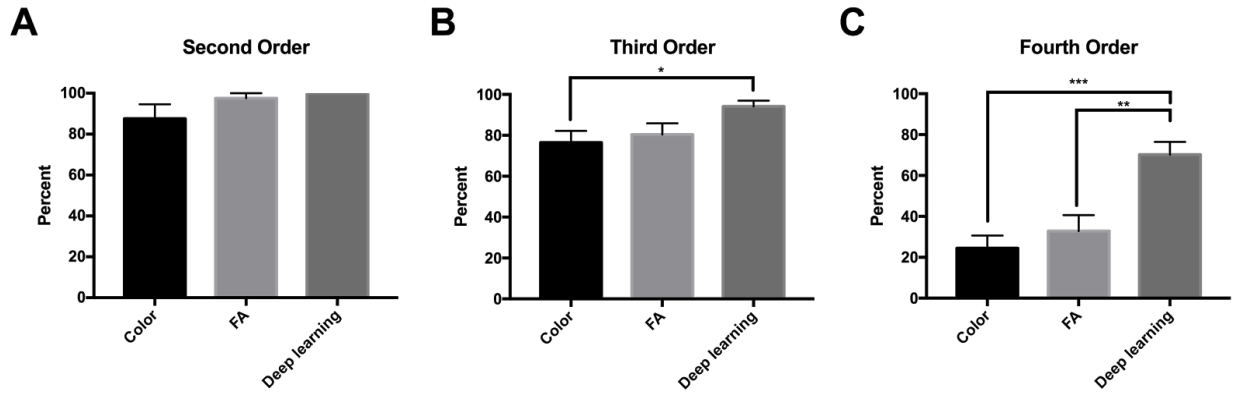

Figure 6. Manual segmentation of vessels on color, fluorescein angiography (FA), and deep learning generated flow maps with OCTA as reference for second order (A), third order (B), and fourth order vessels (C). * P < 0.05, ** P < 0.001, *** P < 0.0001

METHODS

Patients with any retinal diagnoses seen in retina clinics at the University of Washington, Seattle, WA were included in the study. Patients were imaged with both Spectralis OCT (Heidelberg Engineering, Heidelberg, Germany) as standard of care and OCTA as part of the research protocol. Written informed consent was obtained prior to acquisition of OCTA. Patients younger than 18 or non-English speakers were excluded. OCTA images with significant signal quality problems, including signal strength below 6 or excessive motion artifacts were excluded from the study. This study was approved by the Institutional Review Board of the University of Washington and was in adherence with the tenets of the Declaration of Helsinki and the Health Insurance Portability and Accountability Act.

*OCTA imaging*

Participants underwent imaging with a 68 kHz CIRRUS™ HD-OCT 5000 with AngioPlex™ OCT Angiography (ZEISS, Dublin, CA), which operates at a central wavelength of 840nm. To achieve OCTA imaging of retinal vasculature, a repeated B-mode scanning protocol was implemented in the prototype. Four repeated B-scans were acquired at one position and used to extract the blood flow signal as previously described(Q. Zhang et al. 2016) with the total time for single volume acquisition being 3.6 seconds excluding the adjustment time prior to data collection.

In this study, the OCT Angiography system was equipped with motion tracking through an auxiliary real time line scan ophthalmoscope (LSO) and allowed montaging of images with 245 slices obtained per volume.(Q. Zhang et al. 2015) The minimum array for the grid was 3x3, giving a FOV of 6.8 x 6.8 mm$^2$ (approximately 30-40 degrees field of view) while maximum array is 4x6, providing a coverage of 9.0x13.4 mm$^2$. Structural OCT volumes as well as the previously described OCTA algorithm(Wang et al. 2007; Wang 2010; T. Klein et al. 2011) was applied to all the volumetric datasets, and the large en-face OCTA was obtained by stitching the images and maximum intensity projection by use of a software coded with Matlab.(Lin et al. 2016) Structural en-face OCT projections were created in the same manner from the structural OCT volumes.

*Deep learning*

OCTA and OCT volumes taken from diabetic patients were used for training the deep learning algorithm. 70%/15%/15% of volumetric cubes were randomly assigned to training, cross-validation, and testing, respectively. After applying image registration of the OCTA images to the OCT scans, the retinal layers of the OCTA image were segmented and used as ground truth. The input into the model was the unaltered grayscale structural OCT image of the retina. Both the input and the outputs were scaled between 0 and 1. The images were sliced into non-overlapping, vertical strips of 384 px by 128 px to allow deep learning to access the native resolution of the OCT images.

A total of 12 independent deep learning models were tested with varying levels of depth and convolutional filters. Four models were constructed in the style of traditional fully convolutional autoencoder networks with no bridge connection. Four models were constructed with a residual-like element wise summation to combine tensors, and four models were constructed with copy + concatenation to combine tensors in the style of the U-shaped autoencoder (Ronneberger, Fischer, and Brox 2015). (Figure 1A) The activation of the final layer was set to linear activation and Mean Square Error (MSE) was used as the loss function during training. The learning rate was set to 1e-5, and the Adam learning function(Kingma and Ba 2014) was used during training. Batch size was set to 400 images, and the loss was recorded at every training iteration. Dropout was used for the extended training session with the best performing model. Validation MSE was calculated at the end of each epoch and training was stopped when the model began to either overfit or the loss function plateaued. All training was performed using a single server equipped with 8 x Nvidia Tesla P100 graphics processing units with Keras (http://github.com/fchollet/keras), Tensorflow (http://www.tensorflow.org), and NVIDA cuda (v8.0) and cu-dnn (v6.0) libraries (http://www.nvidia.com).

To examine the performance of the trained model, structural OCT volumes from non-diabetic patients with other retinal vasculopathies were used as a held out test set. The cross-sectional structural OCT images from each retina was passed independently through the trained model for inference and projection was performed on the resulting volume. The en-face images from the deep learning output were then compared to the en-face images generated from OCTA.

Comparison against clinicians of ability to identify vessels on structural B scan images was performed by sending four B scan structural images and having three masked retina trained experts segment on the images, utilizing any image enhancement softwares as necessary. The OCTA B scan image was used as ground truth, and the model output was used to compare using paired study design. Sensitivities and specificities were compared using the McNemar test and the positive and negative predictive values were compared using a generalized score statistic.(Leisenring, Alonzo, and Pepe 2000)

Comparison of the model output en-face image against OCTA en- face image, color imaging and fluorescein angiography was performed by registering the four modalities to each other and then randomly sampling 100x100 pixel squares and asking a retina trained clinical expert to segment and label each vessel as second, third, or fourth order vessels. OCTA was used as the reference and the percent visible was calculated for each modality. Fisher exact test was used to calculate significance.